\documentclass[runningheads]{llncs}
\usepackage{graphicx}
\usepackage{xcolor}
\usepackage{amsmath}
\usepackage{amssymb}
\usepackage{bm}
\usepackage{tikz}
\usepackage{pgfplots}
\usetikzlibrary{calc}
\usepackage{leftidx}
\usepackage{subcaption}
\usepackage{siunitx}
\begin{document}
\title{Time-Optimal Transport of Loosely Placed Liquid Filled Cups along Prescribed Paths}
\titlerunning{Task Constrained Time-Optimal Path Following}
%
\author{Klaus Zauner\orcidID{0009-0006-3572-1213} \and Hubert Gattringer\orcidID{0000-0002-8846-9051} \and Andreas M{\"u}ller\orcidID{0000-0001-5033-340X}}
\authorrunning{K. Zauner et al.} 
%
\institute{Institute of Robotics, Johannes Kepler University Linz, \\Altenberger Strasse 69
	4040 Linz, Austria\\
	\email{\{klaus.zauner, hubert.gattringer, a.mueller\}@jku.at}
} 
\maketitle              
\begin{abstract}
Handling loosely placed objects with robotic manipulators is a difficult task from the point of view of trajectory planning and control. This becomes even more challenging when the object to be handled is a container filled with liquid. This paper addresses the task of transporting a liquid-filled cup placed on a tray along a prescribed path in shortest time. The objective is to minimize swapping, thus avoiding spillage of the fluid. To this end, the sloshing dynamics is incorporated into the dynamic model used within the optimal control problem formulation. The optimization problem is solved using a direct multiple shooting approach.


\end{abstract}
\section{Introduction}
Within modern industrial robotics, a pivotal focus lies on movements that are optimized in terms of time. For some applications the movement is to trace a predefined geometric path, parameterized through a designated path parameter. This pursuit of determining the time evolution of the path parameter is commonly known as time-optimized path following \cite{GeuFlores2011}.
In tandem with the inherent limitations imposed by the robotic system, the optimization process must also navigate through constraints linked to the specific task intended during the movement. The scope of this paper is dedicated to the intricate domain of task constraints, with a particular emphasis on resolving the "waiter motion problem" \cite{Flores2013,Gattringer2021}. This problem entails the intricate task of transporting a loosely positioned, liquid-filled cup on a tray affixed to the end-effector of a robotic manipulator \cite{di2021sloshing,di2022sloshing,Yano2001}. The paper commences with the formulation of the general path following problem, expressed in terms of a joint space representation of the path. Subsequently, the formulation systematically extends to encompass task constraints. Initially, constraints are devised to stabilize the position of a general rigid body on the tray, thwarting undesirable movements such as lifting, sliding, and tilting during the motion. These constraints are intricately defined based on the constraining forces exerted on the body. Moving forward, the study advances to consider the fluid dynamics associated with a liquid-filled cup, modeled simplistically through a spherical pendulum \cite{Reinhold2019}. This inclusion of internal dynamics introduces an additional layer of complexity to the optimization problem. The final optimal control problem is solved using a direct multiple shooting method, implemented in the CasADi framework.

\begin{figure}
  \begin{subfigure}{0.59\linewidth}

\begin{tikzpicture}[]
\coordinate (I) at (2,1.5);
\filldraw (I) circle (2pt);
\draw[line width = 0.5mm] (I) circle (2mm);
\node[] at ($(I)+(0.7,0)$) {$\leftidx{_I}{x}{}$};
\draw[-latex, line width = 0.5mm] (I) -- ($(I)+(-1,0)$);
\node[] at ($(I)+(0,1.5)$) {$\leftidx{_I}{y}{}$};
\draw[-latex, line width = 0.5mm] (I) -- ($(I)+(0,1)$);
\node[] at ($(I)+(-1.5,0)$) {$\leftidx{_I}{z}{}$};


\coordinate (H) at (0,0);
\coordinate (B) at (0,-1);
\coordinate (C) at (2,-1.5);
\coordinate (A) at (2.5,-1.5);
\coordinate (EE) at (2.5,-1);
\coordinate (H2) at ($(H)+(5,0)$);
\coordinate (B2) at ($(B)+(5,0)$);
\coordinate (C2) at ($(B2)+(0,-0.5)$);
\coordinate (A2) at ($(C2)+(-1,0)$);
\coordinate (EE2) at (3.75,-1);

\filldraw (H) circle (2pt);
\filldraw (B) circle (2pt);
\filldraw (C) circle (2pt);
\filldraw (A) circle (2pt);
\filldraw (EE) circle (2pt);


\node [] at ($(H)+(0.25,0.25)$) {$H$};

\draw [] (H) -- ($(H)+(0,-0.4)$);
\draw [] ($(H)+(0,-0.6)$) -- (B);
\draw [] ($(H)+(-0.25,-0.4)$) -- ($(H)+(0.25,-0.4)$);
\draw [] ($(H)+(-0.25,-0.6)$) -- ($(H)+(0.25,-0.6)$);


\draw [] (B) -- ($(B)+(0,-0.5)$) -- ($(B)+(0.9,-0.5)$);
\draw [] ($(B)+(1.1,-0.5)$) -- (C);
\draw [] ($(B)+(0.9,-0.75)$) -- ($(B)+(0.9,-0.25)$);
\draw [] ($(B)+(1.1,-0.75)$) -- ($(B)+(1.1,-0.25)$);

\draw [] (C) -- (A);



\node [] at ($(B)+(0.35,0)$) {$B$};

\node [] at ($(C)+(0,-0.35)$) {$C$};

\node [] at ($(A)+(0,-0.35)$) {$A$};

\draw [] ($(EE)+(-0.5,0)$) -- ($(EE)+(0.5,0)$);

\draw [] (A) -- (EE);

\draw [line width=0.5mm, fill=gray!20] ($(EE)+(-0.25,0)$) rectangle ++(0.5,0.75);

\node [] at ($(H)+(-0.45,-0.5)$) {$q_B$};
\node [] at ($(B)+(1,-0.95)$) {$q_C$};

\node [] at ($(EE)+(0.2,-0.2)$) {E};

\node[rotate=-90] at (4.75,0.5) {\includegraphics[width=.5\textwidth]{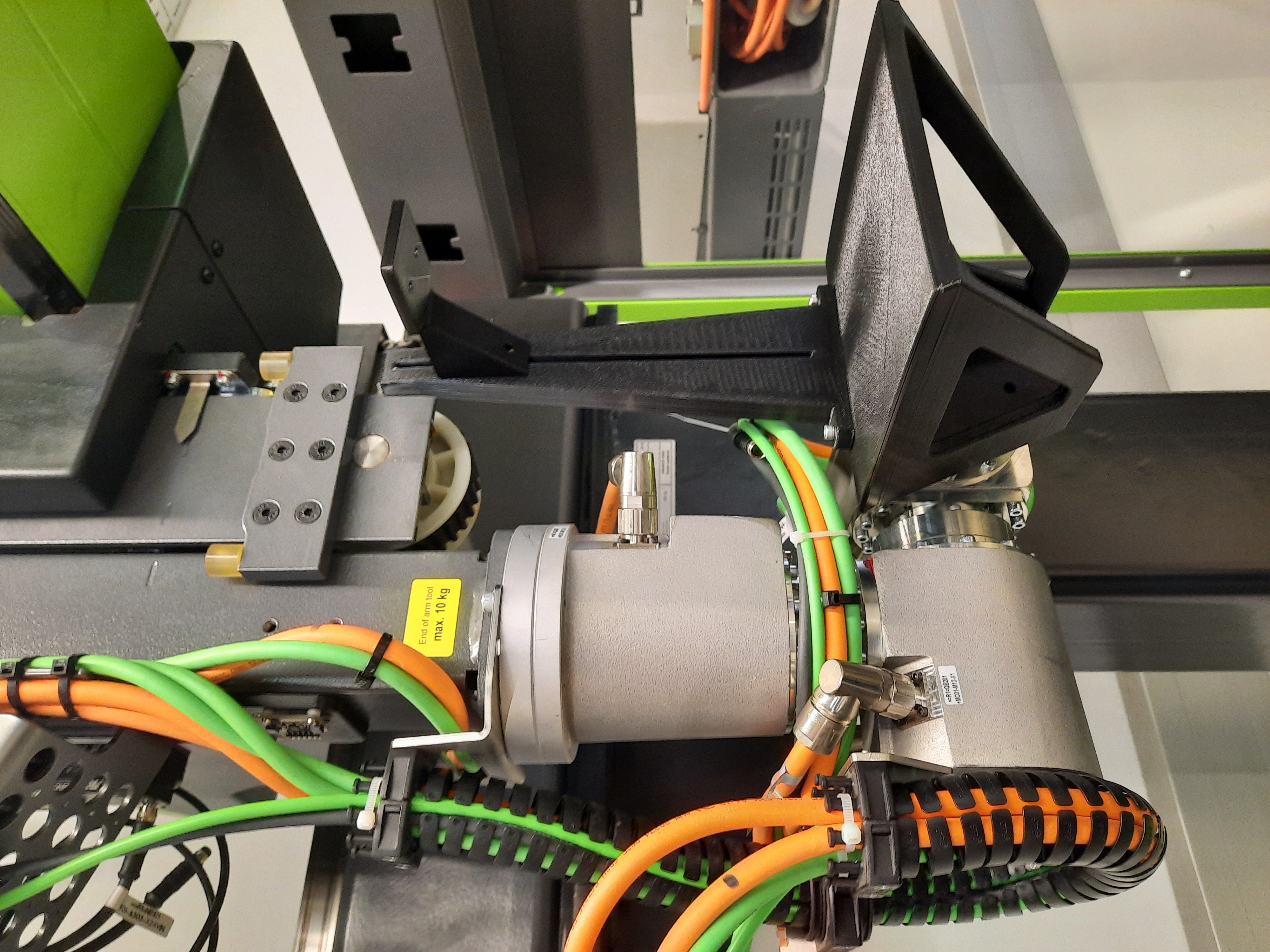}};

\end{tikzpicture}
 \caption{Scheme of the rotational unit with tray and object} \label{fig:Scheme}

  \end{subfigure}%
  \hspace*{\fill}   
  \begin{subfigure}{0.39\linewidth}
	\begin{tikzpicture}[]
	\coordinate (I) at (2,6);

	\coordinate (A) at (2,-1);
	\coordinate (Z) at (2,1.5);
	\coordinate (S) at (2,3);
	\coordinate (K) at (1.5,1.5);
	\draw [line width=0.5mm, fill=gray!20] ($(S)+(-1.5,-1.5)$) rectangle ++(3,3);
	\draw [line width=0.5mm] ($(Z)+(-2,0)$) -- ($(Z)+(2,0)$);

	\filldraw (Z) circle (3pt);

	\draw [-latex, line width=1mm] ($(Z)+(-1.5,0)$) --node[below, near start]{$f_y$} (Z);
	\draw [-latex, line width=1mm] ($(Z)+(0,-1.5)$) --node[left, near start]{$f_z$} (Z);
	\draw [-latex, line width=1mm] ($(Z)+(-45:0.75)$) arc (-45:-250:0.75);
	\node[] at (1.55,2.5) {$M_x$};

	\draw[-latex, line width = 0.5mm] (Z) -- ($(Z)+(1,0)$);
	\node[] at ($(Z)+(0.4,0.8)$) {$\leftidx{_{E}}{z}{}$};
	\draw[-latex, line width = 0.5mm] (Z) -- ($(Z)+(0,1)$);
	\node[] at ($(Z)+(1,0.35)$) {$\leftidx{_{E}}{y}{}$};

	\node [] at ($(Z)+(0.4,-0.3)$) {$\mathcal{F}_E$};

	\end{tikzpicture}
    \caption{Object on tray, $y-z$ view} \label{fig:ObjectContactForces}
  \end{subfigure}%

\caption{Robot scheme (a) and rigid object loosely placed on a tray, mounted on the end effector (b)} \label{fig:1}
\end{figure}
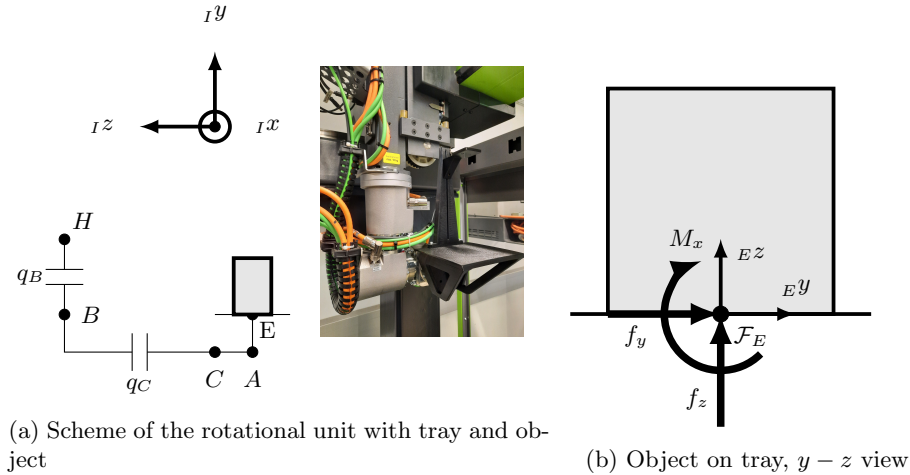
\vspace{-1cm}
\section{Problem Formulation}
\subsection{Time-Optimal Path Following}
Let $\mathcal{Z}\subset\mathbb{R}^6$ denote the task space of a $n$ joint serial robotic manipulator with the end effector position and orientation $\leftidx{_I}{\mathbf{z}}{_{E}^T}=(\leftidx{_I}{\mathbf{r}}{_{E}},\leftidx{_I}{\bm{\varphi}}{_{E}})$, resolved in the inertial frame $\mathcal{F}_I$, and joint coordinates $q_i\in\mathbb{R},\,i\in\{x,y,z,A,B,C\}$. The robot under consideration is basically a three axes linear robot that has been extended by a three axes rotation unit at its end effector. The coupling point is denoted as $H$. This point should be routed along a parameterized path $\leftidx{_I}{\mathbf{r}}{_H}(\sigma):\sigma\in[0,1]\rightarrow\mathbb{R}^3$. The time-optimal path following problem consists in finding the time evolution $\sigma(t),\,t\in[0,t_E]$ such that the $w_t$-weighted end time $t_E$ and the $w_u$-weighted input are minimized and necessary constraints are satisfied. The coupling point $H$ only depends on the coordinates $q_z,\,q_x$ and $q_y$. Therefore we partition the vector $\mathbf{q}=(\mathbf{q}_L^T(\sigma),\mathbf{q}_R^T)^T$ into two parts $\mathbf{q}_L(\sigma)=(q_z(\sigma),q_x(\sigma),q_y(\sigma))^T$ and $\mathbf{q}_R=(q_B,q_C,q_A)^T$. With lower $\underline{(\cdot)}$ and upper $\bar{(\cdot)}$ bounds on the joint velocities $\dot{\mathbf{q}}_L\in\mathbb{R}^3$ and accelerations $\ddot{\mathbf{q}}_L\in\mathbb{R}^3$ the resulting optimization problem is

\begin{equation}
\label{eq:optiPathFollow}
\begin{array}{rc}
 & \underset{t_E,\mathbf{u}}{\text{min}}\,\int\limits_0^{t_E}\left(w_t+w_u\mathbf{u}^T\mathbf{u}\right)\text{d}t\\
\text{s.t.} & \underline{\mathbf{q}}_L\leq\mathbf{q}_L\leq\bar{\mathbf{q}}_L\\
& \underline{\dot{\mathbf{q}}}_L\leq\dot{\mathbf{q}}_L\leq\bar{\dot{\mathbf{q}}}_L\\
& \underline{\ddot{\mathbf{q}}}_L\leq\ddot{\mathbf{q}}_L\leq\bar{\ddot{\mathbf{q}}}_L
\end{array}
\end{equation}


\subsection{Dynamics Model}
\label{subsecDynamics}
The optimization problem \eqref{eq:optiPathFollow} does not account for any dynamics. A dynamic robot model
\begin{equation}
\label{eq:EOMs}
\mathbf{M}(\mathbf{q})\ddot{\mathbf{q}}+\mathbf{G}(\mathbf{q},\dot{\mathbf{q}})\dot{\mathbf{q}}+\mathbf{g}(\mathbf{q})=\mathbf{Q}_{\text{M}},
\end{equation}
with the generalized mass matrix $\mathbf{M}(\mathbf{q})$, the vector of generalized Coriolis and centrifugal forces $\mathbf{G}(\mathbf{q},\dot{\mathbf{q}})\dot{\mathbf{q}}$, gravitational forces $\mathbf{g}(\mathbf{q})$ and the vector of motor torques $\mathbf{Q}_{\text{M}}$, allows to bound the latter ones within the robots capabilities. Taking into account the fact that the first three axes are linear axes and that the rotational degrees of freedom do not have significant arm lengths, the limitation of the motor torques is dispensed and the accelerations are limited instead. As we want to realize the transport of a loosely placed, rigid object along a prescribed path, we need to model the related dynamics. Considering the body, as shown in Fig.\,\ref{fig:ObjectContactForces}, as a subsystem with the describing velocity $\dot{\mathbf{y}}_{\text{obj}}=(\leftidx{_{E}}{\mathbf{v}}{_{E}^T},\leftidx{_{E}}{\bm{\omega}}{_{E}^T})^T$ of the contact point with the tray, the equation of motion results in
\begin{equation}
\mathbf{M}_{\text{obj}}\ddot{\mathbf{y}}_{\text{obj}}+\mathbf{G}_{\text{obj}}\dot{\mathbf{y}}_{\text{obj}}-\mathbf{Q}_{\text{obj}}=\mathbf{Q}_{\text{obj}}^z.
\end{equation}
$\mathbf{Q}_{\text{obj}}^z=\left(\mathbf{f}^{zT},\mathbf{M}^{zT}\right)^T$ denotes the contact wrench with the constraint force $\mathbf{f}^{z}=(f_x,f_y,f_z)^T$ and torque $\mathbf{M}^{z}=(M_x,M_y,M_z)^T$. The contact wrench allows to state the task constraints for the transport, as shown in the following subsection.

\subsection{Task Constraints: Loosely Placed Object}
\label{subsecTaskConstraints}
To ensure successful transportation of the object, it must be prevented from lifting, sliding or tipping over. With the elements of the contact wrench, the related conditions can be formulated as derived in \cite{Gattringer2021}.
\begin{itemize}
\item \textit{Non-lifting condition:} The condition for preventing the loss of contact between object and tray is 
\begin{equation}
\label{eq:constrLift}
f_z\geq0.
\end{equation}

\item \textit{Non-slipping condition:} The slipping of the body is only counteracted by the static friction between the body and the tray. Therefore the dynamic reaction force $||f_T||=\sqrt{f_x^2+f_y^2}$ tangential to the contact plane must be lower than the friction force, or
\begin{equation}
\label{eq:constrSlip}
||f_T||\leq\mu_0f_z.
\end{equation}
The static friction coefficient$\mu_0$ relates to the angle $\rho$ under which sliding start as $\tan{\left(\rho\right)}=\mu_0$. It can therefore be determined via a simple experiment.\\

\item \textit{Non-tipping-over condition:} The torques $M_x$ and $M_y$ or actually the resulting torque would tip the cup around a point on its footprint unless the condition
\begin{equation}
\label{eq:constrTip}
\sqrt{M_x^2+M_y^2}\leq r_of_z,
\end{equation}
with $r_o$ the radius of the cup, is met.

\end{itemize}

\subsection{Task Constraints: Liquid Filled Cup}
\begin{figure}[htb]
\begin{center}
\begin{tikzpicture}[]
	\coordinate (Z) at (0,0);
	\filldraw (Z) circle (3pt);

	\draw[line width = 0.5mm] (-4,0) -- (4,0);
	\draw[line width = 0.5mm] (0,0) -- (3,0) -- (3,5) -- (-3,5) -- (-3,0) -- (0,0);
	\draw[line width = 0.25mm] (0,1.65) -- (0,3);
	\draw[line width = 0.5mm, fill = blue, opacity = 0.2] (0,0) -- (3,0) -- (3,3) -- (-3,3) -- (-3,0) -- (0,0);
	\draw[line width = 0.5mm, fill = blue, opacity = 0.4] (0,0) -- (3,0) -- (3,2) -- (-3,4) -- (-3,0) -- (0,0);
	
	\coordinate (SP) at (0,3);
	\coordinate (m) at ($(SP)+(-5/8,-13/8)$);
	\filldraw (SP) circle (3pt);
	\filldraw (m) circle (10pt);
	\draw[line width = 0.5mm] (SP) -- (m);

	\draw[-latex, line width = 0.5mm] (SP) -- ($(SP)+(1,0)$);
	\node[] at ($(SP)+(0,1.25)$) {$\leftidx{_P}{z}{}$};
	\draw[-latex, line width = 0.5mm] (SP) -- ($(SP)+(0,1)$);
	\node[] at ($(SP)+(1,0.35)$) {$\leftidx{_P}{y}{}$};
	\node [] at ($(SP)+(0.35,-0.35)$) {$\mathcal{F}_P$};

	\draw[-latex, line width = 0.5mm] (Z) -- ($(Z)+(1,0)$);
	\node[] at ($(Z)+(0.4,0.8)$) {$\leftidx{_{E}}{z}{}$};
	\draw[-latex, line width = 0.5mm] (Z) -- ($(Z)+(0,1)$);
	\node[] at ($(Z)+(1,0.35)$) {$\leftidx{_{E}}{y}{}$};


	\draw [-latex, line width=0.25mm] ($(SP)+(-110:1.25)$) arc (-110:-90:1.25);

	\draw [line width=0.5mm] (m) -- ++(1,0);
	\draw [line width=0.5mm] ($(m)+(1,0.2)$) -- ($(m)+(1,-0.2)$);
	\draw [line width=0.5mm] ($(m)+(0.8,0.25)$) -- ($(m)+(1.2,0.25)$) --  ($(m)+(1.2,-0.25)$) -- ($(m)+(0.8,-0.25)$);
	\draw [line width=0.5mm] ($(m)+(1.2,0)$) -- ++(2.4,0);

	\node [] at (0.35,-0.35) {$\mathcal{F}_E$};
	\node [] at ($(m)+(1.5,0.25)$) {$d$};
	\node [] at ($(m)+(-0.5,-0.5)$) {$m$};
	\node [] at ($(SP)+(-100:1)$) {$\varphi$};

	\draw[line width = 0.25mm] (SP) -- ($(SP)+(160:0.75)$);
	\draw[line width = 0.25mm] (m) -- ($(m)+(160:0.75)$);
	\draw[latex-latex, line width = 0.25mm] ($(SP)+(160:0.65)$) -- node[left]{$L$} ($(m)+(160:0.65)$);

\end{tikzpicture}
\caption{Scheme of the tray with a liquid filled cup, y-z view} \label{fig:FluidPendulum}
\end{center}
\end{figure}
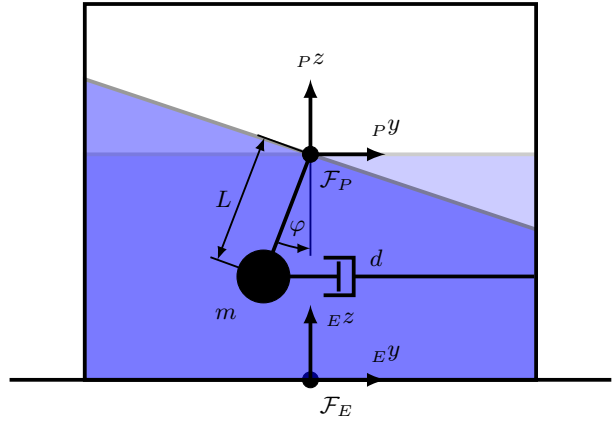

Assuming that the rigid body of section \ref{subsecDynamics} is a liquid filled cup, the fluids dynamics need to be modeled and taken into account during the optimization in order to prevent sloshing. A viable approach is to model the liquid as a spherical pendulum, see Fig.\,\ref{fig:FluidPendulum}, of length $L$ with the fluid representing lumped mass $m$, based on the procedure shown in \cite{Reinhold2019}. The pendulum is modeled as subsystem in the frame $\mathcal{F}_P$ which is aligned with the end effector frame $\mathcal{F}_E$. The describing velocities are $\dot{\mathbf{y}}_{\text{pend}}=(\leftidx{_{P}}{\mathbf{v}}{_{P}^T},\leftidx{_{P}}{\bm{\omega}}{_{P}^T},\dot{\varphi},\dot{\vartheta})^T$. The liquids surface is assumed to stay normal to the pendulum for any pendulum angles $\mathbf{q}_F=(\varphi,\vartheta)^T$, which are identified as angles of a rotation around the $\leftidx{_P}{x}{}$ axis and the resulting $\leftidx{_P'}{y}{}$ axis subsequently. The friction of the wall and the fluids inner damping are modeled with viscous damping $d$. The EOMs are of the form \eqref{eq:EOMs} without the input in form of generalized motor torques. The derivation can be done by using \textit{Lagrangian Equations of the Second Type}. It yields the pendulums dynamics in the general form
\begin{equation}
\label{eq:Pendulum}
\ddot{\mathbf{q}}_F=\mathbf{f}_F(\mathbf{q},\dot{\mathbf{q}},\ddot{\mathbf{q}},\mathbf{q}_F,\dot{\mathbf{q}}_F).
\end{equation}
The most evident condition is that the liquid must not reach the edge of the cup. Again, a state vector $\mathbf{x}_F=(\mathbf{q}_F,\dot{\mathbf{q}}_F)^T$ is introduced, which allows to state the constraints as
\begin{equation}
\label{eq:constrPendulum}
\underline{\mathbf{x}}_F\leq\mathbf{x}_F\leq\bar{\mathbf{x}}_F.
\end{equation}

\section{Solution of the Optimal Control Problem}
The path following for the point $H$ in combination with the liquid transport requires the movement of all six axes of the robot. The movement of the three linear axes is parameterized via the path parameter. In order to obtain continuous accelerations and thus also torques, the jerk is assumed to be piecewise constant as input.
Solving the optimal control problem means the determination of the time evolution of the path parameter $\sigma(t),\,t\in[0,t_E]$ and the required trajectories for the rotational degrees of freedom $\mathbf{q}_R$. Therefore a dynamical system
\begin{equation}
\label{eq:intChain}
\dot{\mathbf{x}}=\mathbf{f}(\mathbf{x},\mathbf{u})
\end{equation}
is formulated where $\mathbf{x}=(\sigma,\dot{\sigma},\ddot{\sigma},\mathbf{x}_R^T,\mathbf{x}_F^T)^T$
denotes the state and $u=(\sigma^{(3)},\mathbf{q}_R^{(3)})$ the input. The corresponding state vectors be $\mathbf{x}_L=(\mathbf{q}_L^T(\sigma),\dot{\mathbf{q}}_L^T(\dot{\sigma}),\ddot{\mathbf{q}}_L^T(\dot{\sigma},\ddot{\sigma}))^T$ and $\mathbf{x}_R=(\mathbf{q}_R^T,\dot{\mathbf{q}}_R^T,\ddot{\mathbf{q}}_R^T)^T$. For the path parameter and the rotatory degrees of freedom, this is a simple integrator chain whereas the dynamics of the state $\mathbf{x}_F$  is given by \eqref{eq:Pendulum}. The optimal control problem (OCP) now states to
\begin{equation}
\label{eq:OCPfull}
\begin{array}{rcr}
& \underset{t_e,\mathbf{u}}{\text{min}}\,\int\limits_0^{t_e}\left(w_t+w_u\mathbf{u}^T\mathbf{u}\right)\text{d}t &\\
\text{s.t.} & \underline{\mathbf{q}}\leq\mathbf{q}\leq\bar{\mathbf{q}}, & -f_z\leq0\\
& \underline{\dot{\mathbf{q}}}\leq\dot{\mathbf{q}}\leq\bar{\dot{\mathbf{q}}}, & \sqrt{f_x^2+f_y^2}-\mu f_z\leq0\\
& \underline{\ddot{\mathbf{q}}}\leq\ddot{\mathbf{q}}\leq\bar{\ddot{\mathbf{q}}}, & \sqrt{M_x^2+M_y^2}-r_of_z\leq0\\
& \underline{\mathbf{x}}_F\leq\mathbf{x}_F\leq\bar{\mathbf{x}}_F &\\
& \dot{\mathbf{x}}=\mathbf{f}(\mathbf{x},\mathbf{u}) &
\end{array}
\end{equation}
Out of the manifold techniques for solving the OCP \eqref{eq:OCPfull} we choose the direct multiple shooting method implemented in the CasADi framework \cite{Andersson2018} using IPOPT \cite{Waechter2006}. The number of shooting intervals is set to $N=400$ and the numerical integration of the cost function and the ODE \eqref{eq:intChain} is performed by using a \textit{Runge-Kutta} scheme of 4th order. The weighting factors are set to $w_t=1$ and $w_u=10^{-4}$.

\section{Results}
The following results are for a lemniscate-path in the horizontal plane,
\begin{equation}
\leftidx{_I}{\mathbf{r}}{_H^T}(\sigma)=\left(\frac{a\sqrt{2}\cos{(\sigma2\pi)}}{\sin{(\sigma2\pi)}^2+1},\frac{a\sqrt{2}\cos{(\sigma2\pi)}\sin{(\sigma2\pi)}}{\sin{(\sigma2\pi)}^2+1},0\right),
\end{equation}
of the point $H$, with $\sigma\in[0,1]$.

\begin{table}[htb]
	\caption{Numerical values of the kinematic parameters, the parameters for the cup transport constraints \eqref{eq:constrLift}, \eqref{eq:constrSlip} and \eqref{eq:constrTip} and the fluid pendulum model.}
	\label{tab:parametersOptimization}
	\begin{center}
		\begin{tabular}{c|c||c|c||c|c}
			\multicolumn{2}{c}{Kinematics} & \multicolumn{2}{c}{Cup} & \multicolumn{2}{c}{Fluid}\\\hline\hline
			$r_{AE,x}$ in $\si{m}$ & 0.07 & $\mu_0$ in $1$ & 0.35 & $L$ in $\si{m}$ & 0.027\\\hline
			$r_{AE,y}$ in $\si{m}$ & 0 & $r_o$ in $\si{m}$ & 0.05 & $m$ in $\si{kg}$ & 0.55\\\hline
			$r_{AE,z}$ in $\si{m}$ & 0.055 & & & $d$ in $\si{\frac{kg}{s}}$ & 0.2\\\hline
			&  & & & $\bar{\mathbf{q}}_F$ in $\si{rad}$ & $\frac{\pi}{18}$\\\hline
			&  & & & $\bar{\dot{\mathbf{q}}}_F$ in $\si{\frac{rad}{s}}$ & $\frac{5\pi}{9}$\\
		\end{tabular}
	\end{center}
\end{table}

\pgfplotsset{
compat=newest,
/pgfplots/myylabel absolute/.style={%
  /pgfplots/every axis y label/.style={at={(0,0.5)},xshift=#1,rotate=90},
  /pgfplots/every y tick scale label/.style={
    at={(0,1)},above right,inner sep=0pt,yshift=0.3em
   }
  }
}

\definecolor{mycolor}{rgb}{0.23935,0.30085,0.54084}%

\definecolor{c1}{rgb}{0,      0.4470,	0.7410}%
\definecolor{c2}{rgb}{0.8500, 0.3250,	0.0980}%
\definecolor{c3}{rgb}{0.9290, 0.6940,	0.1250}%
\definecolor{c4}{rgb}{0.4940, 0.1840,	0.5560}%
\definecolor{c5}{rgb}{0.4660, 0.6740,	0.1880}%
\definecolor{c6}{rgb}{0.6350, 0.0780,	0.1840}%

\def\myWidth{4cm}
\def\myHeight{1.3cm}

\begin{figure}
\centering
\subfloat[Positions of the linear axes]{
\begin{tikzpicture}
\begin{axis}[
	name = plot1,
	width= \myWidth,
	height= 1.5cm,
	scale only axis,
	xmin=0,
	xmax=4.5,
	ymin=-3,
	ymax=3,
	xlabel style={align=center,font=\color{white!15!black}},
	xlabel={$t$ in $s$},
	ylabel style={align=center,font=\color{white!15!black}},
	ylabel={$\mathbf{q}_L$ in $m$},
	myylabel absolute=-35pt,
	axis background/.style={fill=white},
	xmajorgrids,
	ymajorgrids,
	xlabel style={font=\footnotesize},
	ylabel style={align=center,font=\footnotesize},
	ticklabel style={font=\footnotesize}, set layers,
	legend style={
		at={(0.45,1.5)},
		anchor=north,
		font=\footnotesize,
	legend columns=3,
	legend transposed=false},
	]
	\addplot+[solid, color=c1, line cap=round, mark=none, line width=1pt, on layer=axis foreground, restrict x to domain=0:5] table[x=t, y=q_x, col sep=semicolon] {file1.csv};
	\addlegendentry{$q_x$}
	\addplot+[solid, color=c2, line cap=round, mark=none, line width=1pt, on layer=axis foreground, restrict x to domain=0:5] table[x=t, y=q_y, col sep=semicolon] {file1.csv};
	\addlegendentry{$q_y$}
	\addplot+[solid, color=c3, line cap=round, mark=none, line width=1pt, on layer=axis foreground, restrict x to domain=0:5] table[x=t, y=q_z, col sep=semicolon] {file1.csv};
	\addlegendentry{$q_z$}
	
\end{axis}
\end{tikzpicture}
}\hfil
\subfloat[Positions of the rotational axes]{
\begin{tikzpicture}
\begin{axis}[
	name = plot2,
	width= \myWidth,
	height= 1.5cm,
	scale only axis,
	xmin=0,
	xmax=4.5,
	ymin=-3,
	ymax=3,
	xlabel style={align=center,font=\color{white!15!black}},
	xlabel={$t$ in $s$},
	ylabel style={align=center,font=\color{white!15!black}},
	ylabel={$\mathbf{q}_R$ in $rad$},
	myylabel absolute=-35pt,
	axis background/.style={fill=white},
	xmajorgrids,
	ymajorgrids,
	xlabel style={font=\footnotesize},
	ylabel style={align=center,font=\footnotesize},
	ticklabel style={font=\footnotesize}, set layers,
	legend style={
		at={(0.45,1.5)},
		anchor=north,
		font=\footnotesize,
	legend columns=3,
	legend transposed=false},
	]
	\addplot+[solid, color=c1, line cap=round, mark=none, line width=1pt, on layer=axis foreground, restrict x to domain=0:5] table[x=t, y=q_a, col sep=semicolon] {file1.csv};
	\addlegendentry{$q_A$}
	\addplot+[solid, color=c2, line cap=round, mark=none, line width=1pt, on layer=axis foreground, restrict x to domain=0:5] table[x=t, y=q_b, col sep=semicolon] {file1.csv};
	\addlegendentry{$q_B$}
	\addplot+[solid, color=c3, line cap=round, mark=none, line width=1pt, on layer=axis foreground, restrict x to domain=0:5] table[x=t, y=q_c, col sep=semicolon] {file1.csv};
	\addlegendentry{$q_C$}
	
\end{axis}
\end{tikzpicture}
}
\vspace{0.5cm}

\subfloat[Velocities of the linear axes]{
\begin{tikzpicture}
\begin{axis}[
	name = plot3,
	width= \myWidth,
	height= 1.5cm,
	scale only axis,
	xmin=0,
	xmax=4.5,
	ymin=-1.1,
	ymax=1.1,
	xlabel style={align=center,font=\color{white!15!black}},
	xlabel={$t$ in $s$},
	ylabel style={align=center,font=\color{white!15!black}},
	ylabel={$\dot{\mathbf{q}}_L$ normalized},
	myylabel absolute=-35pt,
	axis background/.style={fill=white},
	xmajorgrids,
	ymajorgrids,
	xlabel style={font=\footnotesize},
	ylabel style={align=center,font=\footnotesize},
	ticklabel style={font=\footnotesize}, set layers,
	legend style={
		at={(0.45,1.5)},
		anchor=north,
		font=\footnotesize,
	legend columns=3,
	legend transposed=false},
	]
	\addplot+[solid, color=c1, line cap=round, mark=none, line width=1pt, on layer=axis foreground, restrict x to domain=0:5] table[x=t, y=q_x_d, col sep=semicolon] {file1.csv};
	\addlegendentry{$\dot{q}_x$}
	\addplot+[solid, color=c2, line cap=round, mark=none, line width=1pt, on layer=axis foreground, restrict x to domain=0:5] table[x=t, y=q_y_d, col sep=semicolon] {file1.csv};
	\addlegendentry{$\dot{q}_y$}
	\addplot+[solid, color=c3, line cap=round, mark=none, line width=1pt, on layer=axis foreground, restrict x to domain=0:5] table[x=t, y=q_z_d, col sep=semicolon] {file1.csv};
	\addlegendentry{$\dot{q}_z$}
	\label{eq:coordC}
\end{axis}
\end{tikzpicture}
}\hfil
\subfloat[Velocities of the rotational axes]{
\begin{tikzpicture}
\begin{axis}[
	name = plot4,
	width= \myWidth,
	height= 1.5cm,
	scale only axis,
	xmin=0,
	xmax=4.5,
	ymin=-1.1,
	ymax=1.1,
	xlabel style={align=center,font=\color{white!15!black}},
	xlabel={$t$ in $s$},
	ylabel style={align=center,font=\color{white!15!black}},
	ylabel={$\dot{\mathbf{q}}_R$ normalized},
	myylabel absolute=-35pt,
	axis background/.style={fill=white},
	xmajorgrids,
	ymajorgrids,
	xlabel style={font=\footnotesize},
	ylabel style={align=center,font=\footnotesize},
	ticklabel style={font=\footnotesize}, set layers,
	legend style={
		at={(0.45,1.5)},
		anchor=north,
		font=\footnotesize,
	legend columns=3,
	legend transposed=false},
	]
	\addplot+[solid, color=c1, line cap=round, mark=none, line width=1pt, on layer=axis foreground, restrict x to domain=0:5] table[x=t, y=q_a_d, col sep=semicolon] {file1.csv};
	\addlegendentry{$\dot{q}_A$}
	\addplot+[solid, color=c2, line cap=round, mark=none, line width=1pt, on layer=axis foreground, restrict x to domain=0:5] table[x=t, y=q_b_d, col sep=semicolon] {file1.csv};
	\addlegendentry{$\dot{q}_B$}
	\addplot+[solid, color=c3, line cap=round, mark=none, line width=1pt, on layer=axis foreground, restrict x to domain=0:5] table[x=t, y=q_c_d, col sep=semicolon] {file1.csv};
	\addlegendentry{$\dot{q}_C$}
	
\end{axis}
\end{tikzpicture}
}
\vspace{0.5cm}

\subfloat[Accelerations of the linear axes]{
\begin{tikzpicture}
\begin{axis}[
	name = plot5,
	width= \myWidth,
	height= 1.5cm,
	scale only axis,
	xmin=0,
	xmax=4.5,
	ymin=-1.1,
	ymax=1.1,
	xlabel style={align=center,font=\color{white!15!black}},
	xlabel={$t$ in $s$},
	ylabel style={align=center,font=\color{white!15!black}},
	ylabel={$\ddot{\mathbf{q}}_L$ normalized},
	myylabel absolute=-35pt,
	axis background/.style={fill=white},
	xmajorgrids,
	ymajorgrids,
	xlabel style={font=\footnotesize},
	ylabel style={align=center,font=\footnotesize},
	ticklabel style={font=\footnotesize}, set layers,
	legend style={
		at={(0.45,1.5)},
		anchor=north,
		font=\footnotesize,
	legend columns=3,
	legend transposed=false},
	]
	\addplot+[solid, color=c1, line cap=round, mark=none, line width=1pt, on layer=axis foreground, restrict x to domain=0:5] table[x=t, y=q_x_dd, col sep=semicolon] {file1.csv};
	\addlegendentry{$\ddot{q}_x$}
	\addplot+[solid, color=c2, line cap=round, mark=none, line width=1pt, on layer=axis foreground, restrict x to domain=0:5] table[x=t, y=q_y_dd, col sep=semicolon] {file1.csv};
	\addlegendentry{$\ddot{q}_y$}
	\addplot+[solid, color=c3, line cap=round, mark=none, line width=1pt, on layer=axis foreground, restrict x to domain=0:5] table[x=t, y=q_z_dd, col sep=semicolon] {file1.csv};
	\addlegendentry{$\ddot{q}_z$}
	
\end{axis}
\end{tikzpicture}
}\hfil
\subfloat[Accelerations of the rotational axes]{
\begin{tikzpicture}
\begin{axis}[
	name = plot6,
	width= \myWidth,
	height= 1.5cm,
	scale only axis,
	xmin=0,
	xmax=4.5,
	ymin=-1.1,
	ymax=1.1,
	xlabel style={align=center,font=\color{white!15!black}},
	xlabel={$t$ in $s$},
	ylabel style={align=center,font=\color{white!15!black}},
	ylabel={$\ddot{\mathbf{q}}_R$ normalized},
	myylabel absolute=-35pt,
	axis background/.style={fill=white},
	xmajorgrids,
	ymajorgrids,
	xlabel style={font=\footnotesize},
	ylabel style={align=center,font=\footnotesize},
	ticklabel style={font=\footnotesize}, set layers,
	legend style={
		at={(0.45,1.5)},
		anchor=north,
		font=\footnotesize,
	legend columns=3,
	legend transposed=false},
	]
	\addplot+[solid, color=c1, line cap=round, mark=none, line width=1pt, on layer=axis foreground, restrict x to domain=0:5] table[x=t, y=q_a_dd, col sep=semicolon] {file1.csv};
	\addlegendentry{$\ddot{q}_A$}
	\addplot+[solid, color=c2, line cap=round, mark=none, line width=1pt, on layer=axis foreground, restrict x to domain=0:5] table[x=t, y=q_b_dd, col sep=semicolon] {file1.csv};
	\addlegendentry{$\ddot{q}_B$}
	\addplot+[solid, color=c3, line cap=round, mark=none, line width=1pt, on layer=axis foreground, restrict x to domain=0:5] table[x=t, y=q_c_dd, col sep=semicolon] {file1.csv};
	\addlegendentry{$\ddot{q}_C$}
	\label{eq:coordF}
\end{axis}
\end{tikzpicture}
}
\caption{Time evolution of the robot coordinates and their respective limits}\label{fig:resCoordinates}
\end{figure}
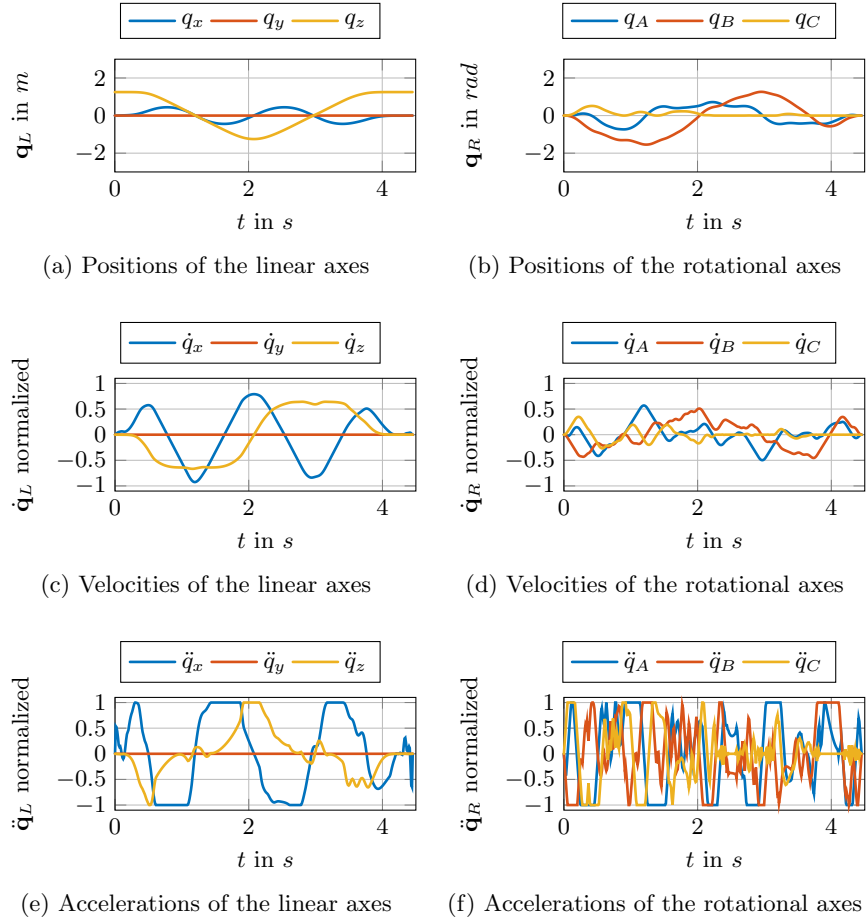

\pgfplotsset{
compat=newest,
/pgfplots/myylabel absolute/.style={%
  /pgfplots/every axis y label/.style={at={(0,0.5)},xshift=#1,rotate=90},
  /pgfplots/every y tick scale label/.style={
    at={(0,1)},above right,inner sep=0pt,yshift=0.3em
   }
  }
}

\definecolor{mycolor}{rgb}{0.23935,0.30085,0.54084}%

\definecolor{c1}{rgb}{0,      0.4470,	0.7410}%
\definecolor{c2}{rgb}{0.8500, 0.3250,	0.0980}%
\definecolor{c3}{rgb}{0.9290, 0.6940,	0.1250}%
\definecolor{c4}{rgb}{0.4940, 0.1840,	0.5560}%
\definecolor{c5}{rgb}{0.4660, 0.6740,	0.1880}%
\definecolor{c6}{rgb}{0.6350, 0.0780,	0.1840}%

\def\myWidth{4cm}
\def\myHeight{1.3cm}

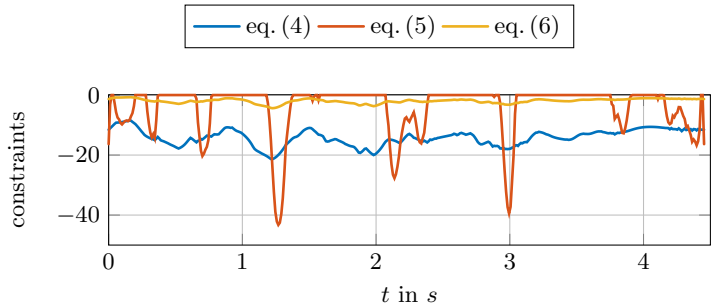
\begin{figure}
\centering
\begin{tikzpicture}
\begin{axis}[
	name = plot1,
	width= 8cm,
	height= 2cm,
	scale only axis,
	xmin=0,
	xmax=4.5,
	ymin=-50,
	ymax=0.1,
	xlabel style={align=center,font=\color{white!15!black}},
	xlabel={$t$ in $s$},
	ylabel style={align=center,font=\color{white!15!black}},
	ylabel={constraints},
	myylabel absolute=-35pt,
	axis background/.style={fill=white},
	xmajorgrids,
	ymajorgrids,
	xlabel style={font=\footnotesize},
	ylabel style={align=center,font=\footnotesize},
	ticklabel style={font=\footnotesize}, set layers,
	legend style={
		at={(0.45,1.6)},
		anchor=north,
		font=\footnotesize,
	legend columns=3,
	legend transposed=false},
	]
	\addplot+[solid, color=c1, line cap=round, mark=none, line width=1pt, on layer=axis foreground, restrict x to domain=0:5] table[x=t, y=cFly, col sep=semicolon] {file1.csv};
	\addlegendentry{eq.\,\eqref{eq:constrLift}}
	\addplot+[solid, color=c2, line cap=round, mark=none, line width=1pt, on layer=axis foreground, restrict x to domain=0:5] table[x=t, y=cSlip, col sep=semicolon] {file1.csv};
	\addlegendentry{eq.\,\eqref{eq:constrSlip}}
	\addplot+[solid, color=c3, line cap=round, mark=none, line width=1pt, on layer=axis foreground, restrict x to domain=0:5] table[x=t, y=cTilt, col sep=semicolon] {file1.csv};
	\addlegendentry{eq.\,\eqref{eq:constrTip}}
\end{axis}
\end{tikzpicture}
\caption{Constraints regarding the losely placed cup}\label{fig:constraints}
\end{figure}

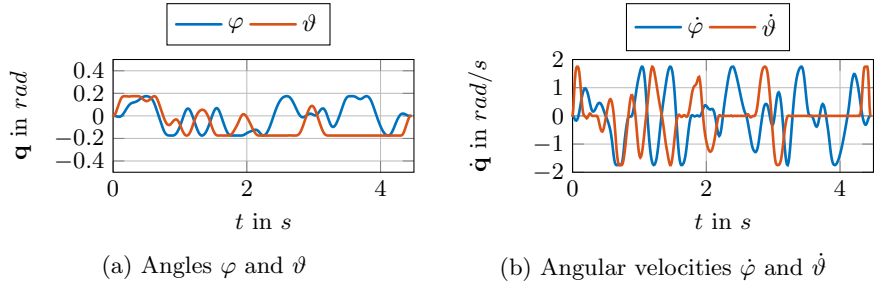
\begin{figure}
\subfloat[Angles $\varphi$ and $\vartheta$]{
\begin{tikzpicture}
\begin{axis}[
	name = plot2,
	width= \myWidth,
	height= 1.5cm,
	scale only axis,
	xmin=0,
	xmax=4.5,
	ymin=-0.5,
	ymax=0.5,
	xlabel style={align=center,font=\color{white!15!black}},
	xlabel={$t$ in $s$},
	ylabel style={align=center,font=\color{white!15!black}},
	ylabel={$\mathbf{q}$ in $rad$},
	myylabel absolute=-35pt,
	axis background/.style={fill=white},
	xmajorgrids,
	ymajorgrids,
	xlabel style={font=\footnotesize},
	ylabel style={align=center,font=\footnotesize},
	ticklabel style={font=\footnotesize}, set layers,
	legend style={
		at={(0.45,1.5)},
		anchor=north,
		font=\footnotesize,
	legend columns=4,
	legend transposed=false},
	]
	\addplot+[solid, color=c1, line cap=round, mark=none, line width=1pt, on layer=axis foreground, restrict x to domain=0:5] table[x=t, y=phi, col sep=semicolon] {file1.csv};
	\addlegendentry{$\varphi$}
	\addplot+[solid, color=c2, line cap=round, mark=none, line width=1pt, on layer=axis foreground, restrict x to domain=0:5] table[x=t, y=theta, col sep=semicolon] {file1.csv};
	\addlegendentry{$\vartheta$}
	
\end{axis}
\end{tikzpicture}
}\hfil
\subfloat[Angular velocities $\dot{\varphi}$ and $\dot{\vartheta}$]{
\begin{tikzpicture}
\begin{axis}[
	name = plot2,
	width= \myWidth,
	height= 1.5cm,
	scale only axis,
	xmin=0,
	xmax=4.5,
	ymin=-2,
	ymax=2,
	xlabel style={align=center,font=\color{white!15!black}},
	xlabel={$t$ in $s$},
	ylabel style={align=center,font=\color{white!15!black}},
	ylabel={$\dot{\mathbf{q}}$ in $rad/s$},
	myylabel absolute=-35pt,
	axis background/.style={fill=white},
	xmajorgrids,
	ymajorgrids,
	xlabel style={font=\footnotesize},
	ylabel style={align=center,font=\footnotesize},
	ticklabel style={font=\footnotesize}, set layers,
	legend style={
		at={(0.45,1.5)},
		anchor=north,
		font=\footnotesize,
	legend columns=4,
	legend transposed=false},
	]
	\addplot+[solid, color=c1, line cap=round, mark=none, line width=1pt, on layer=axis foreground, restrict x to domain=0:5] table[x=t, y=phi_d, col sep=semicolon] {file1.csv};
	\addlegendentry{$\dot{\varphi}$}
	\addplot+[solid, color=c2, line cap=round, mark=none, line width=1pt, on layer=axis foreground, restrict x to domain=0:5] table[x=t, y=theta_d, col sep=semicolon] {file1.csv};
	\addlegendentry{$\dot{\vartheta}$}
\end{axis}
\end{tikzpicture}
}
\caption{Constraints regarding the liquid}\label{fig:constraintsPendulum}
\end{figure}

As Fig.\,\ref{fig:resCoordinates} depicts, the given task can be carried out without any violation of the velocity and acceleration constraints on the robot axes. Figures \ref{eq:coordC} - \ref{eq:coordF} show the joint velocities and accelerations normalized to their respective limits. In Fig.\,\ref{fig:constraints} the constraints \eqref{eq:constrLift}, \eqref{eq:constrSlip} and \eqref{eq:constrTip} are depicted in the form $(\cdot)\leq0$. In order to omit sloshing, the fluid pendulum's angles $\mathbf{q}_F$ and their respective velocities were constrained by \eqref{eq:constrPendulum}. Fig.\,\ref{fig:constraintsPendulum} confirms the compliance of the optimization result. Based on the optimization results, it can be assumed that the liquid transport can be carried out as presented, provided that the assumed parameters are accurate and the liquid model is sufficiently precise.


\section{CONCLUSION}
This paper systematically addresses the challenge of time-optimal path following for an industrial robot, gradually incorporating the complex constraints associated with transporting a loosely positioned, liquid-filled cup. Commencing with the well-established problem of time-optimal path following, our approach involves calculating the constraining forces necessary to maintain the cup's stability during movement. Building upon these forces, we formulate constraints designed to prevent undesired cup behaviors as lifting, sliding, and tilting during the robot's motion. To account for the internal dynamics of the liquid-filled cup, a simplified pendulum model is employed to approximate fluid dynamics. This step aims in constraining the tilt of the liquid surface during the motion. The optimal control problem is solved through the direct multiple shooting method. The choosen target path for the end-effector point $H$ is specified as a lemniscate within the horizontal plane of the workspace.

%

In addition, tests on real robots are planned to verify the procedure. In the future, the model of the liquid-filled cup is also to be extended in such a way that the shift in the center of gravity of the cup due to the movement of the liquid is taken into account when calculating the constraining forces.

\subsubsection*{Acknowledgement}
This work has been supported by the "LCM – K2 Center for Symbiotic Mechatronics" within the framework of the Austrian COMET-K2 program.

%
%
%
 \bibliographystyle{splncs04}
 \bibliography{raad24_bib_zauner}

\end{document}